\documentclass[conference]{IEEEtran}
\IEEEoverridecommandlockouts
\usepackage{multirow}
\usepackage{graphicx}
\usepackage{amsmath,amssymb,amsfonts}
\usepackage{algorithm,algorithmic}
\usepackage[algo2e,linesnumbered]{algorithm2e} 
\usepackage{subfigure}
\usepackage{graphicx}
\usepackage{textcomp}
\usepackage{xcolor}
\usepackage{mathtools}
\usepackage{bbm}
\usepackage{soul}
\usepackage{cite}
\usepackage{makecell}
\usepackage{booktabs}
\usepackage[linkcolor=black,
            anchorcolor=black,
            citecolor=black]{hyperref}
\def\BibTeX{{\rm B\kern-.05em{\sc i\kern-.025em b}\kern-.08em
T\kern-.1667em\lower.7ex\hbox{E}\kern-.125emX}}
\begin{document}

\title{Exploring Extreme Quantization in Spiking Language Models}

\author{\IEEEauthorblockN{Malyaban Bal, Yi Jiang, Abhronil Sengupta}
\IEEEauthorblockA{\textit{School of Electrical Engineering and Computer Science} \\
\textit{The Pennsylvania State University}\\
University Park, PA 16802, USA \\
Email: \{mjb7906, yijiang, sengupta\}@psu.edu}
%\and
%\IEEEauthorblockN{2\textsuperscript{nd} Lu Sen}
%\IEEEauthorblockA{\textit{School of Electrical Engineering and Computer Science} \\
%\textit{The Pennsylvania State University}\\
%University Park, PA 16802, USA \\
%senlu@psu.edu}
%\and
%\IEEEauthorblockN{3\textsuperscript{rd} Abhronil Sengupta}
%\IEEEauthorblockA{\textit{School of Electrical Engineering and Computer Science} \\
%\textit{The Pennsylvania State University}\\
%University Park, PA 16802, USA \\
%sengupta@psu.edu}
}
\maketitle


\begin{abstract}
Despite the growing prevalence of large language model (LLM) architectures, a crucial concern persists regarding their energy and power consumption, which still lags far behind the remarkable energy efficiency of the human brain. Recent strides in spiking language models (LM) and transformer architectures aim to address this concern by harnessing the spiking activity of biological neurons to enhance energy/power efficiency. Doubling down on the principles of model quantization and energy efficiency, this paper proposes the development of a novel binary/ternary (1/1.58-bit) spiking LM architecture. %It employs spikes as neuronal activity and relies solely on 1/1.58-bit synaptic weights, pushing the boundaries of efficiency in LM design. 
Achieving scalability comparable to a deep spiking LM architecture is facilitated by an efficient knowledge distillation technique, wherein knowledge from a non-spiking full-precision ``teacher'' model is transferred to an extremely weight quantized spiking ``student'' LM. Our proposed model represents a significant advancement as the first-of-its-kind 1/1.58-bit spiking LM, and its performance is rigorously evaluated on multiple text classification tasks of the GLUE benchmark.
\end{abstract}


\section{Introduction}

Large language models (LLMs) are becoming increasingly popular due to their wide-ranging applications in various natural language processing (NLP) tasks. The immense scale of transformer \cite{vaswani2017attention} based language models (LMs) comes at the cost of increased energy/power consumption. By integrating bio-plausible models with extreme parameter quantization techniques, we can markedly diminish energy and power costs, thereby fostering a sustainable future for deep neural models.

% Recent breakthroughs in multi-modal foundation models have paved the way for cutting-edge models like GPT-4 and Llama 3, ushering in a new era of multi-modal learning with unprecedented possibilities.

Spiking neural networks (SNNs) \cite{ghosh2009spiking} emulate biological processes, where communication between neurons takes the form of spikes. This sparse, event-driven flow of information enables efficient computation and communication in neuromorphic hardware, leading to significant energy savings \cite{sengupta2019going}. Though substantial work on SNN-based transformer architectures are done on vision-based datasets, recent works have explored spiking LM architectures for language generation \cite{zhu2023spikegpt} and classification tasks \cite{bal2024spikingbert}.

While spiking architectures assist in reducing multi-bit neuron activations to binary spiking activity, they offer limited support in minimizing parameter overhead. Model quantization techniques stands out as a promising solution, effectively slashing the memory footprint and computational expenses of expansive models, all while upholding their competitive performance. Model quantization can occur during training or post-training. While post-training quantization \cite{zhang2023post} is simpler, it often leads to more significant accuracy degradation compared to techniques that incorporate quantization-aware training. Our work is primarily geared towards developing a quantization-aware training approach. 

The spiking architectures' capacity to distribute computations across the temporal domain is pivotal in overcoming the formidable challenge of crafting an exceptionally quantized spiking LM. By amalgamating scalable spiking LM with extreme model quantization \cite{wang2023bitnet, liu2022bit}, this endeavor strives to markedly diminish model size and enhance energy/power efficiency through the utilization of spiking activation and 1/1.58-bit weights. This quantized spiking architecture is adaptable for implementation on neuromorphic chips and specialized hardware accelerators such as ``In-Memory" binary neural network (BNN) accelerators \cite{lu2020exploring}. Moreover, executing the binarized spiking LM over an optimal number of time-steps enables us to reach near full-precision accuracies while capitalizing on the specialized ``In-Memory'' hardware accelerator tailored for BNNs \cite{lu2020exploring}.

In our paper, we introduce a framework for training a spiking LM with parameters quantized to 1/1.58-bits. We leverage the technique of implicit differentiation at equilibrium for training the spiking architecture, as outlined in previous works \cite{bal2024spikingbert, xiao2021training}. This approach offers outstanding memory efficiency during training compared to Backpropagation Through Time (BPTT), which demands substantial memory for storing a large computational graph. Moreover, it eliminates the need for surrogate gradient methods by implicitly computing gradients, thus addressing the non-differentiability challenge inherent in training spiking models with BPTT. Within specific constraints \cite{bai2019deep}, this learning paradigm parallels biologically plausible and energy-efficient training methods, such as equilibrium propagation \cite{scellier2017equilibrium, bal2022sequence, lin2024scaling}, thereby reinforcing a neuromorphic perspective on learning.

In this paper, leveraging the average spiking rate at equilibrium of the quantized spiking LM, we use an efficient knowledge distillation (KD) technique to transfer knowledge from a non-spiking high-precision ``teacher'' architecture to a spiking 1/1.58-bit ``student'' LM. This KD methodology is pivotal for the creation of an extremely quantized spiking LM, enabling efficient training of our ``student'' model even with limited resources.
For this study, we adopt the encoder-based BERT \cite{devlin2018bert} architecture as our LM, chosen for its suitability in text classification tasks due to its ability to capture bi-directional contextual information.

The primary contributions of this paper are as follows: 
\begin{itemize}
\item We propose an efficient framework for training-aware extreme model quantization in spiking LM architectures.

  \item We leverage the equilibrium dynamics of the spiking LM to efficiently perform the model compression and quantization through the use of KD.
 
  \item To the best of our knowledge, this is the first-of-its-kind 1/1.58-bit spiking LM which is evaluated on multiple NLP tasks of the GLUE benchmark.
  
\end{itemize}

\section{Methods}

In this section, we begin by exploring the architecture of our 1/1.58-bit BERT-based spiking LM. Subsequently, we provide a concise overview of the training mechanism and the effective KD technique employed, enabling us to achieve such extreme quantization by distilling knowledge from a full-precision ``teacher'' model.

\subsection{Architecture and Learning Dynamics}

\begin{figure}[h]
\centering
\label{classdis}
\includegraphics[width=.7\columnwidth]{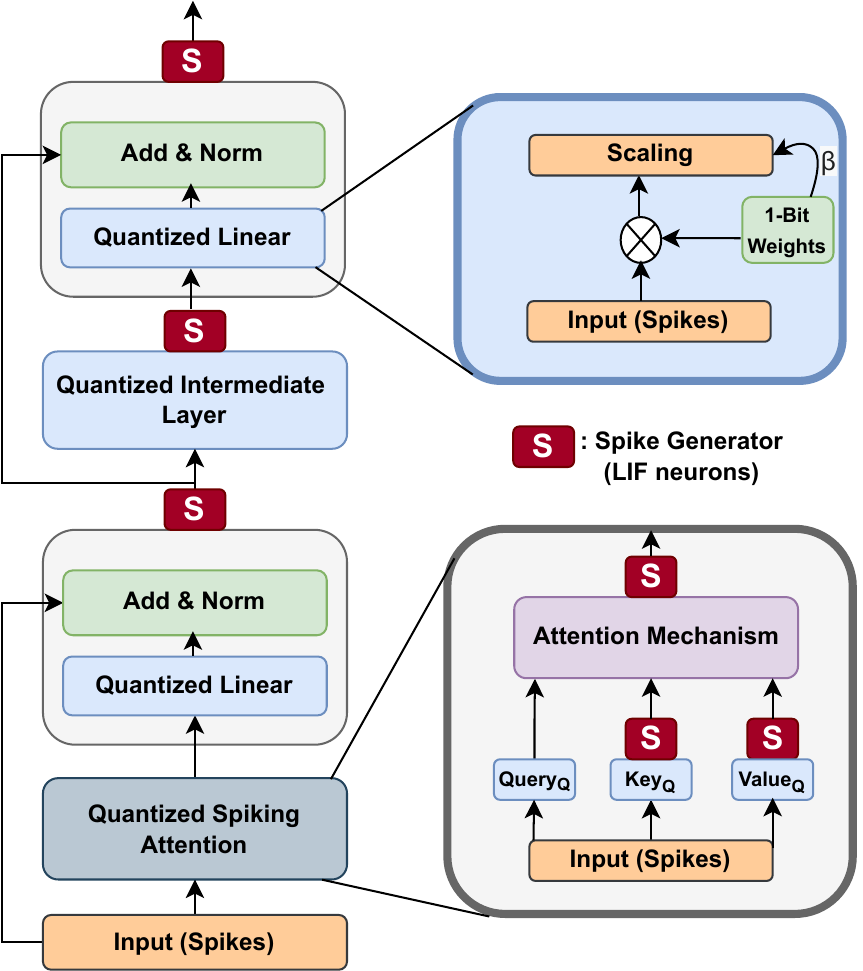}
\caption{High-level architecture of each encoder layer of the 1-bit Spiking transformer architecture with Quantized Linear layers. All linear layers, including those used in the attention module, are quantized to 1-bit.}
\label{fig1}
\end{figure}

The base architecture of this model follows previous BERT-based spiking architectures \cite{bal2024spikingbert}, and comprises of stacked spiking encoder layers. Each spiking encoder layer comprises of a spiking attention module and multiple intermediate layers as described in Fig \ref{fig1}. In this highly quantized version, we substitute all linear projection layers, typically full precision feed-forward layers, with quantized binary/ternary linear layers, whose operation is detailed later. Additionally, given the spiking nature of this architecture, communication between neuronal layers exclusively takes place in the form of spikes (Fig. \ref{fig1}). The dynamics of the spiking neurons are given as:

\begin{equation}
\label{eqn1}
\begin{aligned}
u_i[t + \delta] = \gamma  u_i[t] + \sum_j(w_{ij}s_j[t]) + b_i, \\
s_i[t+1] = S(u_i[t + \delta]), \\
u_i[t + 1] = u_i[t + \delta] - V_{th} s_i[t+1]
\end{aligned}
\end{equation}
Here, $\gamma$ denotes the leaky term; $V_{th}$ is the spiking threshold; $u_i[t]$ and $s_i[t]$ signifies the membrane potential and spike from the $i^{th}$ neuron at time $t$; $w_{ij}$ represents the synaptic weight between the pre and post-synaptic neurons; $t+\delta$ denotes an intermediate time step to determine if the neuron has fired; $b_i$ indicates a bias term; $S$ denotes the non-differentiable function for spike generation, with subtraction as reset operation.

The average spiking rate (ASR) for each layer $i$, which can be defined as a weighted-average function as follows, $a_i[t] =  \frac{\sum^t_{\tau = 1}\gamma^{t-\tau}s_i[\tau]}{\sum^t_{\tau = 1}\gamma^{t-\tau}}$,  is leveraged during training and during KD following \cite{bal2024spikingbert} and is given as,
\begin{equation}
\label{eqn2}
\begin{aligned}
a_{i}[t+1] = \frac{1}{V_{th}}(W_{i}(a_{(i-1)}[t+1]) + b_i - \frac{u_{i}[t+1]}{\sum^t_{i=0}\gamma^i})
\end{aligned}
\end{equation}
For linear layers, Eqn. \ref{eqn2} can be derived from Eqn. \ref{eqn1}. Theoretically, as $t\rightarrow \infty$ the ASRs eventually converge to equilibrium and we can derive steady-state equations for ASR of linear layers and formulate surrogate steady-state functions for non-linear layers \cite{xiao2021training, bal2024spikingbert} of the form:
\begin{equation}
\begin{aligned}
\label{eqn3}
a_i^* = \sigma (\frac{1}{V_{th}}(W_{i}(a^*_{i-1}) + b_i))
\end{aligned}
\end{equation}
where clipping function $\sigma(x)$ bounds the values within [0,1] and $W_i$ is operation of the corresponding layer. The layer-wise convergence dynamics of ASRs of the 1-bit quantized Spiking LM model is shown in Fig. \ref{fig2}a. We leverage only the ASR values at equilibrium to compute error-gradients using implicit differentiation at equilibrium \cite{xiao2021training} as shown below, 
\begin{equation}
\label{eqn4}
\frac{\partial L(a^*)}{\partial \theta} = - \frac{\partial L(a^*)}{\partial a^*} (J^{-1}_{g_{\theta}}|_{a^*}) \frac{\partial f_{\theta}(a^*)}{\partial \theta}
\end{equation}
where, $L$ is the loss function used, $g_{\theta}(a) = f_{\theta}(a) - a$, $f$ is the steady-state equation of ASR, $J^{-1}$ is the inverse Jacobian of $g_\theta$ when $a = a^{*}$, i.e., at equilibrium. Hence, unlike BPTT, we eliminate the need to store all the intermediate states. Additionally, we circumvent the need for surrogate gradients to address the non-differentiability issue encountered by BPTT during SNN training, as we employ ASRs (real values) rather than spikes during training. This capacity to utilize ASRs during training is also crucial for formulating an extremely quantized spiking architecture, as detailed later.

\subsection{Quantizing linear layers to 1/1.58-bit}

The predominant energy expenditure during inference of a spiking LM arises from the numerous linear projection layers utilized across various components, including the attention module and intermediate layers such as feed-forward and output layers. This is primarily attributed to the floating-point matrix multiplication operation performed by these layers, which incurs significantly higher energy costs compared to normalization or residual layers \cite{wang2023bitnet}. Linear projection operations involve two tensors: neuronal activity and synaptic weights. Primarily, previous research has typically concentrated on extreme quantization ($< 2$ bits) of either neuronal activity \cite{bal2024spikingbert} or synaptic weights \cite{wang2023bitnet, bai2022towards}. In contrast, our objective is to simultaneously quantize both using just 1 bit (neurons are intrinsically quantized due to spiking behavior). We further explore ternary weight quantization, utilizing a $log_2(3)$ i.e. 1.58-bit representation, where the states correspond to -1/0/1, in order to bridge the accuracy gap with the full-precision model even further.

\begin{figure}
  \centering
  \includegraphics[width=\columnwidth]{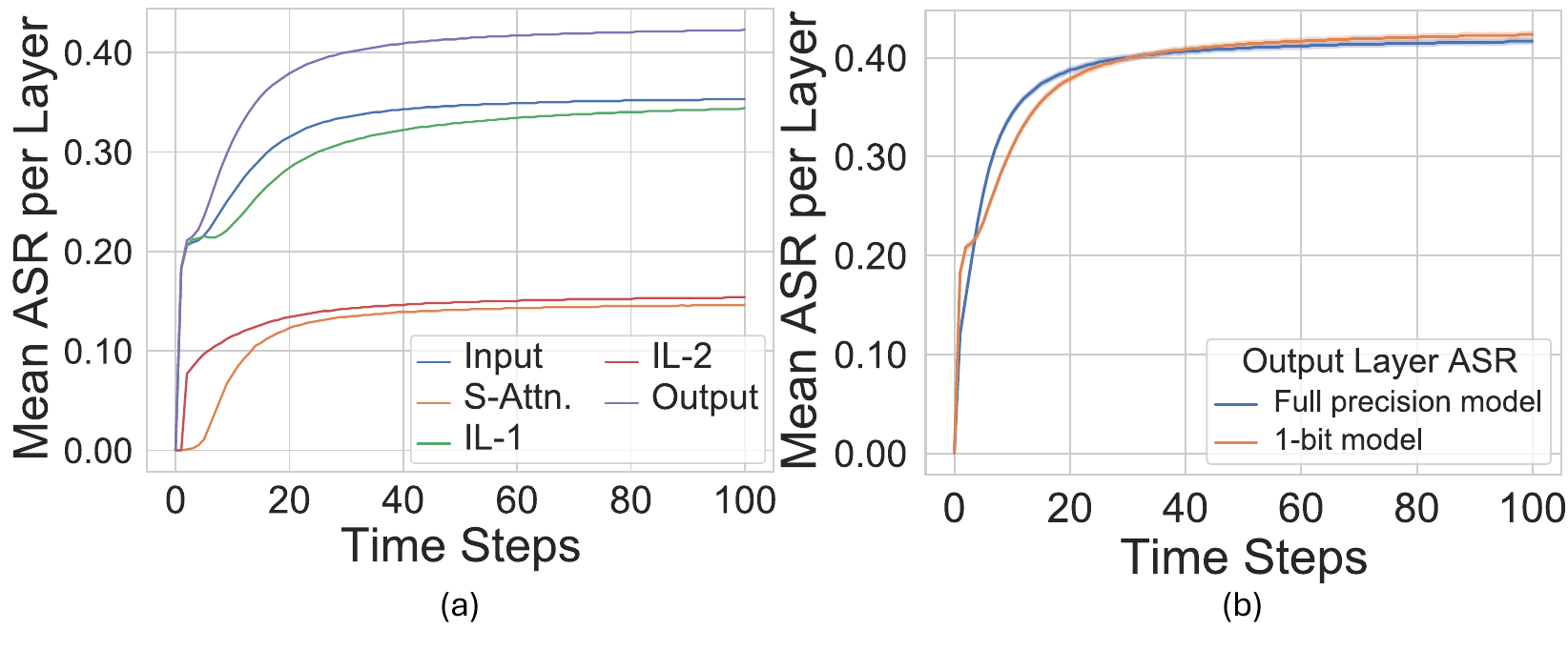}
  \caption{Results obtained after passing a set of randomly sampled inputs from MRPC dataset. (a) The convergence dynamics of the different sub-layers of an encoder layer of the 1-bit SpikingBERT\textsubscript{4} model. (b) Comparison of output layer ASR convergence dynamics of full precision (weights) and 1-bit SpikingBERT\textsubscript{4}. Y-axis in both shows  mean (over number of neurons) of the ASR while X-axis shows the operating time steps. }
  \label{fig2}
\end{figure}

In line with \cite{wang2023bitnet, liu2022bit}, for 1-bit quantization, the weights ($W$) are first centred around zero-mean and then the  model weights are binarized to either +1 or -1 using the signum function. Thus, after binarization the quantized weight is, 
\begin{equation}
\begin{aligned}
\label{eqn5}
W_{Q_{1-bit}} = Sign(W-\alpha)
\end{aligned}
\end{equation}
% \begin{equation}
% \label{eqn6}
% \text{Sign}(W_{ij}) = \begin{cases}
% +1, & \text{if } W_{ij} > 0, \\
% -1, & \text{if } W_{ij} \leq 0.
% \end{cases}
% \end{equation}
where, $\text{Sign}(W_{ij}) = \begin{cases}
+1, & \text{if } W_{ij} > 0, \\
-1, & \text{if } W_{ij} \leq 0.
\end{cases}$ and $\alpha = \frac{1}{\text{nm}} \sum_{ij} W_{ij}
$ is the mean of the weight matrix $W \in \mathbb{R}^{n \times m}$. For the 1.58-bit implementation, the quantization can be done as follows:
\begin{equation}
\begin{aligned}
\label{eqn6}
W_{Q_{1.58-bit}} = Round(Clip(W/(\beta + \epsilon), -1, 1) 
\end{aligned}
\end{equation}
where, $\epsilon$ is a small constant. The output of the linear layer is scaled by a constant $\beta = \frac{1}{\text{nm}} \sum_{ij} |W_{ij}|
$. In addition to the extreme quantization, our spiking LM distinguishes itself from non-spiking designs by quantizing the input to linear layers, solely by binary spikes, contrasted with continuous values \cite{ma2024era, bai2022towards, wang2023bitnet}. It is crucial to emphasize that we harness this spiking activity during inference. During training, as detailed in the previous section, we utilize the full-precision ASR which helps us maintain the theoretical constraints of quantization \cite{wang2023bitnet}. Even with extreme quantization, comparable performance to full-precision model is attributed to the fact that the spiking LM operates over $T$ timesteps, effectively spreading neuron activation precision across the temporal dimension. Additionally, as illustrated in Fig. \ref{fig2}b, the ASR convergence dynamics of the quantized model aligns closely with that of the full-precision model, indicating no additional overhead on the operating latency of the spiking LM.

\subsection{Knowledge Distillation (KD)}

Training large spiking transformer models, such as those found in LMs, demand significant resources. Our approach, which depends on training-aware extreme quantization, exacerbates the challenge of training from scratch. Building on prior research \cite{bal2024spikingbert} regarding KD from ANN to SNN models, our goal is to streamline the training process. We achieve this by employing a full-precision ``teacher'' LM to impart knowledge to a quantized spiking ``student'' LM. This approach enables efficient compression of knowledge and significantly reduces the required resources compared to training from scratch. This formulation, akin to Eqn. \ref{eqn4}, capitalizes on the steady-state ASR of the output of intermediate layers of the quantized spiking LM. It utilizes this information to minimize a loss function, incorporating the full precision activity of corresponding intermediate layers within a ``teacher''. This mechanism facilitates the effective transfer of knowledge leveraging the intermediate layers—specifically, the output of each individual encoder block—of both the ``teacher'' and the ``student'' model. The underlying loss is given as,
\begin{equation}
\begin{aligned}
\label{eqn7}
{L}_{h_i} = MSE(ASR(S_{h_i}^*)W_{p}, T_{f(h_i)})
\end{aligned}
\end{equation}
Here, $MSE$ is mean squared error loss function, $ASR(S^*_{h_i})$ represents the equilibrium ASR of the output neurons in the $i^{th}$ SE layer of the ``student'' model, while $T_{f(h_i)}$ denotes the output of the $f(h_i)^{th}$ layer in the ``teacher'' model. $W_{p}$ signifies a linear projection aligning the dimensionality of the ``student'' layer with its corresponding layer in the ``teacher''. The function $f$ maps the ``student'' layer $h_i$ to a specific target layer in the ``teacher'' network.

\section{Results}

In this section, we showcase the efficacy of extreme quantization techniques primarily on text classification tasks of the General Language Understanding Evaluation (GLUE) benchmark \cite{wang2018glue}. The experiments were conducted on Nvidia RTX A5000 GPUs (8 units), each equipped with 24GB of memory.

\subsection{Datasets}
We used Quora Question Pair (QQP), Microsoft Research Paraphrase Corpus (MRPC) to evaluate our model's performance on similarity and paraphrase tasks. For inference-oriented evaluations, we opted for Multi-Genre Natural Language Inference (MNLI), Question-answering NLI (QNLI) datasets. For single-sentence sentiment analysis tasks, we utilized the Stanford Sentiment Treebank (SST-2). For all tasks, we keep the maximum sequence length at 128. 

\subsection{Experimental Setup}
All experiments  use four 1/1.58-bit spiking encoder blocks (Fig. \ref{fig1}) where encoding dimension of the tokens in the input is 768 and the intermediate size of the model is 3072. Training starts with a full-precision BERT-based spiking LM \cite{bal2024spikingbert}, pre-trained on the Wikipedia corpus. Subsequently, we conduct task-specific internal layer KD for each individual dataset. To quantize the weights, we employ Eqn. \ref{eqn5} for 1-bit and Eqn. \ref{eqn6} for 1.58-bit variant, followed by intermediate layer KD as per Eqn. \ref{eqn7} with non-spiking full-precision fine-tuned BERT as ``teacher''. After KD, we refine our quantized models by minimizing cross-entropy loss against the true labels. This facilitates the training of our quantized models tailored to the specific task. Since this is the first work of exploring extreme quantization in spiking LM, our results are compared with some common NLP models and other efficient implementation of BERT. The results are demonstrated in Table. \ref{table1}. The model trained with ternary (1.58-bit) weights outperform that trained with binary (1-bit) weights and closely match the performance of full-precision SpikingBERT. %By reducing the previous 32-bit weights of a similarly scaled spiking architecture to just 1/1.58 bit, we achieve substantial memory savings. 
Internal-layer KD (Eqn. \ref{eqn7}) is crucial for high accuracy in quantized models. Without it, quantization-aware training on the proposed spiking LM leads to a significant 6-8\% accuracy drop across the datasets.

\begin{table}
\centering
\footnotesize
\caption{
Results Showing Performance (Accuracy) of Our Quantized Spiking Models  Against Some Standard Models and Other Efficient Implementations of BERT on GLUE Evaluation Set.
}
\begin{tabular}{llllll}
\hline
\textbf{Model} & \textbf{QQP} & \textbf{MNLI} & \textbf{SST-2} & \textbf{QNLI} &  \textbf{MRPC} \\
\hline
CBoW \cite{wang2018glue} & 75.0 & 57.1 & 79.5 &  62.5 &  75.0\\
REM W2-A4 \cite{bai2022towards} & 75.7 & 58.3  & 82.9 & 75.3 & - \\
BinaryBERT\textsubscript{$50\%$} \cite{qin2022bibert} & 66.7 & 39.2  & 54.1 & 59.5 & 68.3 \\
TernaryBERT \cite{qin2022bibert} & 74.1 & 32.7  & 53.1 & 59.3 & 68.3 \\
BERT\textsubscript{5} + PF \cite{xu2021bert} & 84.1 & 67.7 & 81.6 & 80.9 & 78.6 \\
NAS-BERT\textsubscript{5} + KD \cite{xu2021bert} & 85.8 & 74.4  & 87.3 & 84.9 & 79.6 \\
BERT\textsubscript{TINY} Adam \cite{bal2024spikingbert} & 81.1 & 65.3  & 80.1 & 77.8 & 69.9 \\
\textbf{1-bit SpikingBERT\textsubscript{4}} & \textbf{83.8} &  \textbf{75.4} &  \textbf{86.7} &  \textbf{80.5} &  \textbf{75.8}\\
\textbf{1.58-bit SpikingBERT\textsubscript{4}} & \textbf{85.4} &  \textbf{77.1} &  \textbf{87.1} &  \textbf{83.1} &  \textbf{77.3}\\
\hline
SpikingBERT\textsubscript{4} & 86.8 & 78.1 & 88.2 & 85.2 &  79.2\\
\end{tabular}

\label{table1}
\end{table}

\subsection{Energy \& Power Efficiency}

%In this study, we implement profound weight quantization, converting all linear layers, which typically account for the majority of energy consumption as network scale increases \cite{wang2023bitnet}, to solely utilize integer accumulation (ACC) operations rather than floating-point multiplicative and accumulative (MAC) operations. In the context of 45nm CMOS technology \cite{han2015learning}, integer ACC operations exhibit 46 times greater energy efficiency compared to floating-point MAC operations, as demonstrated by the energy consumption ratio of $4.6 \text{pJ} / 0.1 \text{pJ}$. 
As demonstrated empirically in Fig. \ref{fig2}b for the output layer, the spiking activity of neurons post-quantization remains comparable to that of  full-precision models. 
Furthermore, empirical results suggest that the total number of $Norm\#OPS$ in full-precision (floating point ACC) and 1/1.58-bit models (Integer ACC) are similar; for example, for the MRPC dataset, the ratio of $Norm\#OPS$ of 1.58-bit SpikingBERT to full-precision SpikingBERT is 1.06 (i.e., $\approx 1$). The total normalized OPS \cite{lu2020exploring} is defined as $Norm\#OPS = \frac{\sum_i{IFR_i*Layer\#OPS_{i+1}}}{\sum{Layer\#OPS}}$, where $IFR_i$ represents the number of spikes over inference time steps averaged across the number of neurons. However, it is worth noting here that each accumulative operation in 1/1.58-bit quantized models are at least an order of magnitude more energy efficient than full-precision models \cite{lu2020exploring} (for instance, a preliminary estimate in terms of integer ACC versus floating point ACC energy consumption in 45nm CMOS technology yields $9\times$ energy efficiency \cite{han2015learning}). Thus, the combination of extreme weight quantization and spiking neuronal activity enables a remarkable reduction in the model's size, energy and power consumption.

\section{Conclusion \& Future Works}

In our paper, we delve into leveraging the equilibrium-based convergence dynamics of spiking architectures to craft an extremely quantized spiking LM. This approach drastically shrinks model size, facilitating deployment on edge-based computing devices with limited resources. 
Furthermore, using computationally efficient ``In-Memory'' BNN accelerators over time enables us to seamlessly perform inference time tradeoff of accuracy and energy in complex sequence-based NLP tasks.
% Moreover, employing 1-bit synaptic weights and neuronal activations streamlines floating-point MAC operations in matrix multiplication to simpler integer ACC and also allows us to leverage custom ``In memory'' hardware accelerators tailored for BNNs, leading to substantial energy and power savings. 

Future endeavors can broaden the application of this technique to encompass other deep spiking architectures, including GPT-like decoder-based LMs. Additional research efforts can aim to minimize the accuracy disparity between the full-precision models and quantized spiking transformer-based architectures. This technique of extreme quantization can also be explored in other deep spiking architectures beyond LMs.

\section*{Acknowledgments}
This material is based upon work supported in part by the U.S. National Science Foundation under award No. CAREER \#2337646, CCSS \#2333881, CCF \#1955815, and EFRI BRAID \#2318101 and by Oracle Cloud credits and related resources provided by the Oracle for Research program.

%\bibliography{cites}
%\bibliographystyle{IEEEtran}


\end{document}